# Using Syntactic Features for Phishing Detection

Gilchan Park
CIT
Purdue University
West Lafayette, IN USA
park550@purdue.edu

Julia M. Taylor
CIT & CERIAS
Purdue University
West Lafayette, IN USA
jtaylor1@purdue.edu

*Abstract*—This paper reports on the comparison of the subject and object of verbs in their usage between phishing emails and legitimate emails. The purpose of this research is to explore whether the syntactic structures and subjects and objects of verbs can be distinguishable features for phishing detection. To achieve the objective, we have conducted two series of experiments: the syntactic similarity for sentences, and the subject and object of verb comparison. The results of the experiments indicated that both features can be used for some verbs, but more work has to be done for others.

*Keywords—phishing detection; syntactic similarity; parse tree path.*

## I. INTRODUCTION

This paper aims to report the comparison between phishing emails and legitimate emails in terms of syntactic similarity and subjects and objects of verb phrases for sentences. To accomplish our goal, two series of experiments were performed with three data sets consisting of two phishing corpora and a legitimate corpus. The phishing corpora is comprised of old phishing emails collected in 2005 [1] and up-to-date phishing emails reported in 2014 [2]. The two phishing corpora were used to observe whether the pattern in phishing emails have changed over time, with respect to subject and object of the verbs; and, if the difference exists, how significant it is.

We compare a corpus of regular everyday emails [3] with phishing emails to emulate human experience in respect to what kind of emails they receive. We have conducted the experiments using the list of target verbs frequently appearing in phishing emails. The syntactic similarity analysis had a purpose of comparing paths from a target verb to the other nodes in syntax trees generated from phishing sentences and legitimate sentences. The subsequent experiment was to examine the difference in subjects and objects of target verbs between the two corpora. For the purpose of this paper, a subject is defined as the noun phrase which serves as the subject of a verb within a segment of a sentence, such as a clause. An object is referred as a noun or a pronoun which is the head of the syntactic object of the verb. For instance, in sentence *We need you to confirm your identity in order to regain full privileges of your account.* we define the subject of the verb *'confirm'* as *you* and the object of the verb *'confirm'* is *identity*. Our primary interest was to investigate whether those two features could work as distinguishable characteristics for phishing emails.

In the previous research, we have analyzed phishing emails based on keyword matching [4]. This paper suggests an approach that goes beyond simple word comparison by considering elements of sentence structure as well as other constituents of the target verb in a sentence. In other words, the scope of the analysis unit has expanded from words to sentence segments. For our purposes, we hypothesized that a verb might bring up different elements in a sentence and also might lead to a disparate syntactic structure of a sentence depending on the intention of usage of the verb. For instance, phishers may want to use the word *update* in an attempt to gain personal information such as *update your account*, on the other hand, 'normal' users probably choose update for other purposes in email as well, such as in *I'll update the document*. The experimental results, however, indicated that such differences could not be generalized as the distinguishable features for all verbs in phishing and legitimate emails but were rather verb specific. The syntactic formation and syntactic components for a sentence have revealed the limitations on identifying phishing scams from unlabeled emails. One of the reasons may be the difference in a number of meanings (senses) that each verb may have, not detectable by syntactic analysis.

The ultimate goal of our research is to extract robust features to be generalized in order to discriminate between phishing emails and legitimate emails. In an attempt to discover any specific patterns related to syntactic features of phishing attempts, we presented a textual syntactic analysis in this paper. The proposed approach solely depends on syntactic aspects of sentences. In the future research, we will investigate semantic characteristics of sentences for phishing detection.

## II. BACKGROUND

This section will briefly review the prior works on phishing detection techniques and pertinent information on syntactic similarity relative to the parse tree path. The previous studies for the parse tree path will be followed.

### A. Prior Phishing Detection Techniques

With the increased attention to phishing, various approaches of phishing detection have been proposed. The approaches vary in their results, but all seem to be successful according to their metrics. The earlier methods were based on blacklisting [5], [6], [7], and [8], or link and URL analysis [9], [10], and [11]. Both mechanisms are still quite popular and widely-deployed in the industry. However, the applications on

This research was partially funded by a seed grant from the College of Technology of Purdue University.

the basis of those techniques revealed the limitations since the phishers continue to change the malicious websites and adopt an auto-generated system for URLs. In addition, one of the main drawbacks of those applications was to falsely flag legitimate sources as phishing.

Later attempts to identify phishing concentrates on contents of emails. For example, Hajgude and Ragha [12] and Xiang et al. [13] proposed phishing detection approaches using characteristics extracted from the contents. In particular, Verma et al. [14], Lee et al. [15], Park & Taylor [16], Park [4] focused on identifying whether the emails contained phishing attempts by analyzing the natural language features in text of the emails.

*B. Parse Tree Path*

For general natural language text, Gildea and Jurafsky [17] first introduced the parse tree paths for their automatic semantic role labeling system. They used the FrameNet database, where semantic roles are defined as frames and the essential components of frames are called frame elements [18]. For example, the frame 'judgment' has 'judge', 'evaluee', 'reason', and 'role' frame elements. The following sentence is the example stored in the FrameNet database: [$_{Judge}$ She ] blames [$_{Evaluee}$ the Government ] [$_{Reason}$ for failing to do enough to help ]. Since frame elements were annotated by humans, as a way of automatically annotating frame elements, the authors created the parse tree paths. The parse tree paths are the routes following the sentence's syntax tree from the target word to other possible frame elements tagging up ( ↑ ) or down ( ↓ ) annotation. The figure 1 below is an example of the parse tree path. Gildea and Jurafsky calculated the probability distributions for generated parse tree paths in order to identify frame element boundaries. For instance, the path ↑ VB ↑ VP ↓ NP had a high probability in their training data, and thereby the direct object of the target verb will play as a frame element.

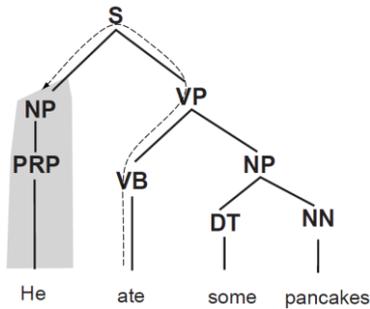

Fig. 1. The parse tree path from the frame element 'He' to the target word 'ate' is represented as ↑ VB ↑ VP ↑ S ↓ NP (see [17])

Swanson and Gordon [19] compared five alternatives for encoding parse tree paths. Their aim was to identify which of them could generate the most accurate parse tree path of a sentence in terms of the accuracy of their argument labeling corresponding to the sentence's annotated semantic role information. The feature of parse tree path is heavily related to the syntactic structure produced by the parser, and therefore using different parsers creates different parse tree paths. In particular, the researchers compared traditional constituency parsers and dependency parsers. For the constituency parsers, they used the Charniak parser [20] and Stanford parser [21], and for the dependency parsers, three variations of Minipar parsers [22]. The results of their experiment indicated that the constituency parsers in overall performed better than the dependency parser.

Gordon and Swanson [23] adopted the feature of parse tree paths in order to identify syntactically similar verbs for an automated semantic role labeling algorithm. Their premise was that verbs appearing in syntactically similar sentences would be very likely to have analogous relations with their arguments. In other words, they used the parse tree paths for measuring the syntactic similarity of pairs of verbs. For instance, their top pairs of verbs based on syntactic similarity include plunge:tumble, dip:fell, pluck:whisk, dive:jump.

Some drawbacks of the algorithm were compensated for by Sagae and Gordon [24]. Containing part-of-speech tags in the encoding of the parse tree paths was one of the drawbacks since it brought about zero results from the cosine calculation between verbs of different classes. Sagae and Gordon removed the parts of speech tags of target verbs, and substituted the tags of end nodes with a generic terminal label. In addition, as a way of differentiating same paths having different end nodes based on the original parse tree path rules, the authors added four more tags ( ↖ ↗ ↘ ↙ ) for the direction of the transition. For instance, the path described in the figure 1 is represented as ↗ VP ↖ S ↙ NP by their feature representation.

The research in this paper adapted the parse tree path feature to compare the syntactic similarity of verbs. In particular, the main focus of the comparison was to examine whether the target verbs had dissimilar syntactic structures between phishing and legitimate emails. We applied the Stanford parser to create parse tree paths, and tagged on all nodes in a path to closely investigate the difference in syntactic structures for the same verbs between phishing and legitimate emails.

### III. EXPERIMENTAL SETUP

Two main experiments were conducted. The first experiment was designed to investigate the sentence's syntactic similarity by the target verbs between phishing corpus and legitimate corpus. The subsequent experiment aimed to compare subjects and objects of the target verbs between those two different sets of sentences.

For the experiments, two data sets were prepared: phishing and legitimate emails. The phishing email set consists of two different corpora, referred here as the Nazario corpus and the APWG corpus. The Nazario corpus was taken from a publicly available collection of phishing emails[1], with 4558 emails, and the APWG corpus was constructed from the emails provided by Anti-Phishing Working Group [2]. The APWG phishing corpus consist of emails that are reported by users to the group. In this experiment we used phishing emails reported in September 2014, containing 80384 emails. The legitimate emails are the sample emails from the public Enron email set by the CALO Project[3].

The Nazario phishing corpus and the legitimate corpus were first used in the rule-based phishing email detection approach [4], where the original emails were preprocessed in order to extract text body. Since we have analyzed sentences in these experiments, the data sets were preprocessed, removing unnecessary information, such as headers, forwarded text, etc. Some of the preprocessed data sets contained duplicate emails and non-English emails in case of phishing corpus, which were removed. We followed the same process for the APWG corpus. After eliminating duplicates, the size of phishing corpus was 2856 emails for Nazario and 25706 emails for APWG corpora,, and the size of legitimate corpus was 3828 emails. We follow the most frequent verb list reported in [4] for this experiment, resulting in the verbs: *access, click, confirm, enter, follow, protect, update, use*. The Figure 2 below describe the overall process to generate similarities.

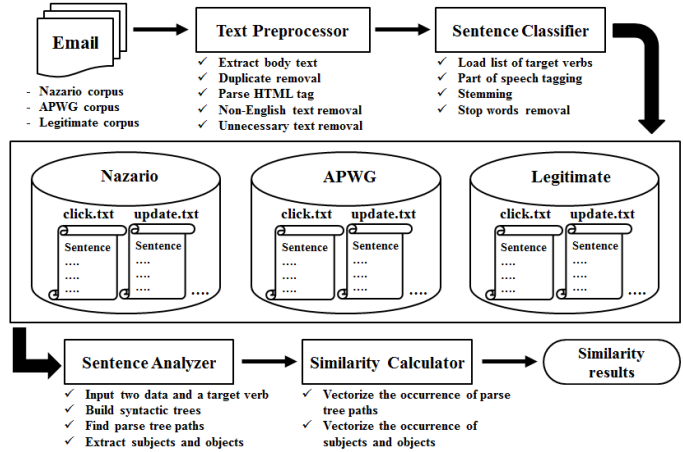

Fig. 2. The process of similarity analysis.

## A. Syntactic similarity

The Stanford Part of Speech Tagger version 3.4.1 and Stanford Parser version 3.4.1 was used to create sentences' syntactic trees. Once the sentences were represented as syntactic trees, the parse tree paths were found starting from the verb to the other nodes, and the frequency of parse tree paths was counted. As an example, table I demonstrates the top 3 parse tree paths for the verb *update* from each dataset with the example sentences.

TABLE I.  THE TOP 3 PARSE TREE PATHS OF THE VERB 'UPDATE'

| Rank | Corpus | The most frequent Parse Tree Path & example |
|---|---|---|
| 1 | Phishing Nazario | ↑VB↑VP↓NP↓PRP$ <br> [ update → your ] : To update your eBay records click here. |
| 1 | Phishing APWG | ↑VB↑VP↓NP↓NN <br> [ update → account] : You need to update your account before the link expires , after 24 hours. |
| 1 | Legitimate | ↑VB↑VP↓NP↓NN <br> [ update → business ] : These people should update their respective business units. |
| 2 | Phishing Nazario | ↑VB↑VP↓NP↓NNS <br> [ update → records ] : Please update your records on or before July 10, 2005. |
| 2 | Phishing APWG | ↑VB↑VP↓NP↓PRP$ <br> [ update → your ] : Please update your information within 72 hours. |
| 2 | Legitimate | ↑VB↑VP↑VP↓TO <br> [ update → to ] : today can I grab you for a few minutes after the close to update you on fund stuff. |
| 3 | Phishing Nazario | ↑VB↑VP↓NP↓NN <br> [ update → account ] : Please follow the link below to update your account information. |
| 3 | Phishing APWG | ↑VB↑VP↑VP↓TO <br> [ update → to ] : You are required to update through the link below. |
| 3 | Legitimate | ↑VB↑VP↓NP↓PRP$ <br> [ update → my ] : Can I update all my Netscape Bookmarks at the same time ? |

In order to measure the similarity for two parse tree path sets from phishing corpus and legitimate corpus, cosine similarity coefficient was used. For each verb, two vectors (a phishing vector and a legitimate vector) used for cosine similarity contained all the parse tree paths from both phishing corpus and legitimate corpus as the components. The occurrences of the parse tree paths in the phishing corpus and legitimate corpus were the values of components for phishing vector and legitimate vector respectively. For instance, if the parse tree path ↑VB↑VP↓NP↓PRP$ appears 1331times for the verb ′update′ in the APWG phishing corpus, and 27 times in legitimate corpus, the value in the phishing vector is 1331for the path ↑VB↑VP↓NP↓PRP$, and the value in the legitimate vector is 27. The cosine similarity scores for the parse tree paths are listed in table II.

TABLE II.  THE COSINE SIMILARITY FOR PARSE TREE PATHS BETWEEN LEGITIMATE AND CORRESPONDING PHISHING DATA

| Verb | Cosine similarity | |
|---|---|---|
| | Nazario | APWG |
| access | 0.2923 | 0.2234 |
| click | 0.5865 | 0.6944 |
| confirm | 0.4388 | 0.5094 |
| enter | 0.5279 | 0.5503 |
| follow | 0.269 | 0.4184 |
| protect | 0.4196 | 0.5936 |
| update | 0.5547 | 0.6729 |
| use | 0.5789 | 0.7028 |

The first column presents the similarities between Nazario phishing data and legitimate data, and the second column presents the similarities between APWG phishing data and legitimate data. It was expected that a verb would construct a different syntactic tree based on its overall intent, but the results indicated that the syntactic structures were not drastically different between the phishing emails and legitimate emails. As the table demonstrates, the range in similarity is greater in the APWG comparison relative to the Nazario one.

All verbs but *access* had the higher similarity scores when the APWG corpus was used than Nazario corpus was used; however, the overall increase was not noticeable enough to consider since the maximum score is just above 70%. In addition, the similarity values vary depending on the verbs. For example, the verbs *access* and *follow* had much lower scores than the other verbs in the first column, and the score gap between *access* and *use* was almost 0.5 for the APWG comparison. This suggested that the difference in the parse tree paths was strongly affected by verbs themselves. To sum up, the syntactic structure of a sentence was not enough to be a distinguishable feature for phishing emails (for all verbs) because of the insufficient and inconsistent similarity scores.

*B. Subject and Object similarity*

The next experiment compared the subjects and the objects of target verbs between the sentences in the phishing and legitimate corpora. The underlying assumption of this experiment was that scammers and normal users might use the same verbs for different purposes. This experiment aimed to investigate whether the disparate purpose of verbs caused the difference in subjects and objects of the verbs, and if the difference existed, whether the difference could be generalized enough to distinguish between phishing emails and legitimate emails. The subject and object similarity measurements were also performed between the two phishing corpora in order to investigate how much they differ.

When it comes to finding the subject and the object of a verb in a sentence segment, the Stanford typed dependencies representation (SD) [25] was adopted. SD represents the simple description of grammatical relations in a sentence as binary relations. The binary relations indicate the grammatical relations between the dominating constituent such as a verb and the dependent or dominated constituent such as a subject or an object of the verb. The Stanford Parser (version 3.4.1) provides SDs.

*1) Subject comparison*

Table III describes the most frequent subject of each target verb, its frequency, and an example sentence.

TABLE III. THE MOST FREQUENT SUBJECT OF TARGET VERBS

| Verb | Corpus | The most frequent subject (percentage) & example |
|---|---|---|
| confirm | Phishing Nazario | **you** (85.26%) *We need you to confirm your identity in order to regain full privileges of your account.* |
| confirm | Phishing APWG | **you** (53.79%) *You must confirm your Paypal account before we close it.* |
| confirm | Legitimate | **I** (13.46%) *I just want to confirm the trades I have in your book.* |
| update | Phishing Nazario | **you** (75.58%) *That requires you to update the order Information.* |
| update | Phishing APWG | **you** (71.84%) *You are required to update through the link below.* |
| update | Legitimate | **you** (43.60%) *From there you will be able to update your email information securely.* |
| follow | Phishing Nazario | **you** (44.68%) *To do so we need you to follow the link below and proceed to confirm your information.* |
| follow | Phishing APWG | **you** (46.15%) *You are kindly advised to follow the instructions below to re-instate your accounts..* |
| follow | Legitimate | **I** (12.78%) *I will follow up with a phone call.* |
| access | Phishing Nazario | **you** (79.85%) *After your verification process is completed you will be able to access your account again.* |
| access | Phishing APWG | **they** (55.28%) *If someone else has access to your account, they have your password and might be trying to access your personal information or send junk email.* |
| access | Legitimate | **you** (77.78%) *You can easily access these real-time features through Yahoo!* |
| click | Phishing Nazario | **you** (85.71%) *You must click the link to complete the process.* |
| click | Phishing APWG | **you** (45.23%) *If you can't click the above link, move this email to your inbox and then click!* |
| click | Legitimate | **you** (60.98%) *If you click on a link in this email and it doesn't open properly in your browser, try copying and pasting the link directly into your browser's address or location field.* |
| enter | Phishing Nazario | **you** (86.19%) *PayPal will never ask you to enter your password in an email.* |
| enter | Phishing APWG | **you** (44.91%) *You will be prompted to enter a new password.* |
| enter | Legitimate | **you** (29.82%) *You will then be asked to enter your destination.* |
| protect | Phishing Nazario | **you** (65.64%) *To ensure your Visa card's security, it is important that you protect your Visa card online with a personal password.* |
| protect | Phishing APWG | **you** (16.86%) *You have been requested to verify your identity and protect your online account.* |
| protect | Legitimate | **efforts**[1] (9.52%) *Robin Kapiloff, an analyst at Moody's Investors Service, said the city 's efforts to diversify its economy over the past decade will protect its revenue collections, even as some of the city 's biggest employers suffer.* |
| use | Phishing Nazario | **you** (56.98%) *If you use popup killers please disable them.* |
| use | Phishing APWG | **you** (39.61%) *You will not be able to use your HSBC account with us until it has been reactivated.* |
| use | Legitimate | **you** (17.22%) *You cannot use Netscape.* |

[1] The word, while the most frequent one, appeared only twice with this verb. The rest of the subjects appeared only once.

As can be seen from the table, the most frequent subjects between the two corpora were quite similar. However, the distribution of the subjects indicated that the most frequent subjects in phishing emails were more dominant than those in legitimate emails. The raw numbers may not do justice to similarity in subjects. For example, in some cases, such as in *click*, APWG phishing corpus contained more imperative sentences (*click here*) compared to the Nazario corpus. Although the subject of imperative sentences like that is implied *you*, the parser could not infer the subject. Considering the imperative sentences, the distribution of the subject *you* for *click* will be still dominant. This fits with the hypothesis that the verbs in phishing emails are used in very particular ways, while in legitimate emails they are used to articulate different needs, and for varied purposes.

*2) Object comparison*

This section compared the object of the target verbs. The most frequent object of a target verb with an example sentence is described in the table IV.

TABLE IV. THE MOST FREQUENT OBJECT OF TARGET VERBS

| Verb | Corpus | The most frequent object (percentage) & example |
|---|---|---|
| confirm | Phishing Nazario | **identity** (31.35%) *Please confirm your identity here.* |
| | Phishing APWG | **information** (25.08%) *We need to confirm your account information..* |
| | Legitimate | **meeting** (18%) *Phillip, This message is to confirm our meeting with you on.* |
| update | Phishing Nazario | **records** (47.79%) *Please update your records in maximum 24 hours.* |
| | Phishing APWG | **information** (25.93%) *Please update your information within 72 hours.* |
| | Legitimate | **profile** (17.24%) *If you're not signed in, you will need to do so before you can update your profile.* |
| follow | Phishing Nazario | **link** (61.48%) *To confirm your identity please follow the link below.* |
| | Phishing APWG | **us** (21.94%) *Follow us on Twitter.* |
| | Legitimate | **instructions** (7.14%) *If you no longer wish to receive Autoweb.com's monthly newsletter, please follow the instructions below.* |
| access | Phishing Nazario | **form** (55.25%) *Please click the hyperlink below to access the Regions InterAct Customer Form.* |
| | Phishing APWG | **information** (45.73%) *If someone else has access to your account, they have your password and might be trying to access your personal information or send junk email.* |
| | Legitimate | **Email/Calendar** (19.51%) *Does anyone have permission to access your Email/Calendar?* |
| click | Phishing Nazario | **link** (55.69%) *To begin upgrading your account please click the link below.* |
| | Phishing APWG | **link** (40.78%) *Please click the link below.* |
| | Legitimate | **link** (43.69%) *Should you choose to not be contacted at this email address again, please click this link and enter the email address you wish to have removed.* |
| enter | Phishing Nazario | **password** (42.67%) *PayPal will never ask you to enter your password in an email.* |
| | Phishing APWG | **password** (21.94%) *On the next screen, enter a new password of your choice.* |
| | Legitimate | **symbols** (21.79%) *Enter multiple symbols separated by a space.* |
| protect | Phishing Nazario | **account** (34.11%) *Thanks for your patience as we work together to protect your account.* |
| | Phishing APWG | **you** (20.99%) *Please understand that this is a security measure intended to help protect you and your account.* |
| | Legitimate | **PC** (13.79%) *Robert tells you what's at stake and how to protect your PC.* |
| use | Phishing Nazario | **account** (16.31%) *In accordance with NCUA User Agreement, you can use your online account in 24 hours after activation.* |
| | Phishing APWG | **links** (10.39%) *Please use links below for details.* |
| | Legitimate | **it** (5.41%) *I do not want to ask for interest free money if Enron will not use it.* |

Unlike the subjects, the most frequent objects between the two corpora were all different except for the verb *click*. The percentage of the most objects were not as high compared to the subjects in phishing emails. It is still significantly higher than those in legitimate emails due to the fact that numerous sentences were analogous in phishing emails.

We also measured the overall cosine similarity for the subjects and objects. For the cosine scores of the subjects, two vectors from phishing and legitimate emails for each verb were considered. The occurrences of subjects in each dataset were the values of the vectors. The cosine scores of the objects were calculated the same way. The results of cosine similarities are shown in table V. The overall result indicates that the subjects for most verbs are quite similar, but the objects are different between the two corpora.

The results are consistent with the results obtained for the most frequent verbs. Interestingly, the verbs *confirm* and *protect* had a much lower score in subject similarity than the other verbs, and the verb *click* had a significantly greater score in object similarity than the other verbs for both phishing corpora. This shows that the similarity seemed to depend on verbs again.

TABLE V. THE COSINE SIMILARITY FOR SUBJECTS AND OBJECTS BETWEEN LEGITIMATE AND CORRESPONDING PHISHING DATA

| Verb | Cosine similarity | | | |
|---|---|---|---|---|
| | Subject | | Object | |
| | Nazario | APWG | Nazario | APWG |
| access | 0.9868 | 0.544 | 0.0733 | 0.0241 |
| click | 0.9824 | 0.9652 | 0.9066 | 0.9003 |
| confirm | 0.2433 | 0.3402 | 0.0153 | 0.0513 |
| enter | 0.8712 | 0.883 | 0.227 | 0.2133 |
| follow | 0.6489 | 0.6555 | 0.22 | 0.3162 |
| protect | 0.001 | 0.0485 | 0.0724 | 0.1715 |
| update | 0.8769 | 0.8953 | 0.2316 | 0.4152 |
| use | 0.7364 | 0.8345 | 0.2372 | 0.4229 |

The table VI reports the cosine similarities between the two phishing corpora. It should be reminded that while the email duplicates were removed, sentence duplicates from various emails were not. All verbs except *access* had similar subjects. The overall similarity scores of objects were relatively high compared to those in legitimate corpus. The low score for *access* was due to a dominance of several sentences in each corpus: *Please select the hyperlink and visit the address listed to access the form.* (179 out of total 324 sentences in Nazario) *If someone else has access to your account, they have your password and might be trying to access your personal information or send junk email.* (330 out of total 597 sentences in APWG) The sentence in APWG was also the reason that the subject score was relatively low compared to the other verbs' scores. In case of *use*, the object score was low as well. Due to its nature of usage, the verb *use* appeared in many different sentences. The objects for *use* were the most sparsely distributed in those for the other verbs.

TABLE VI. THE COSINE SIMILARITY FOR SUBJECTS AND OBJECTS BETWEEN NAZARIO AND APWG

| Verb | Cosine similarity | |
|---|---|---|
| | Subject | Object |
| access | 0.5458 | 0.2479 |
| click | 0.9788 | 0.8945 |
| confirm | 0.9808 | 0.8941 |
| enter | 0.9649 | 0.8329 |
| follow | 0.8911 | 0.6518 |
| protect | 0.6941 | 0.7616 |
| update | 0.9856 | 0.6138 |
| use | 0.9222 | 0.3274 |

IV. CONCLUSION

In this paper, we presented the comparison of sentence syntactic similarity and the difference in subjects and objects of target verbs between phishing emails and legitimate emails. For the syntactic similarity, we explored the parse tree paths for the verbs. The result showed that the parse tree paths were not 100% distinguishable between phishing emails and legitimate emails, but the difference was high enough. Looking at the result, most scores ranged between 40 to 70 percent that indicating that syntactic structure similarity depends on verbs themselves. This suggests that the syntactic structures of sentences driven by verbs are not enough to play a role in a definite differentiating between phishing and legitimate categories. Most of the verbs, however, have multiple meanings. We will explore the effect of meaning on the path structure as well as classification in our future work.

The results of the subject comparison show that the most frequent subject of the verbs was the personal pronoun *you* in both phishing and legitimate emails. Intuitively, it is possibly predictable that the subject of the verbs in phishing emails can be *you* since the attackers' primary purpose is to ask the recipient to perform some actions. The interesting result was that the most frequent subject of verbs in legitimate emails was *you* as well. One possible explanation for this is that the target verbs were selected based on the frequent verbs in phishing emails, and the chosen verbs have a common general usage, namely to require recipients to do something. Given that assumption, the scope of subjects for the target verbs might be limited. The cosine similarity showed that the overall subjects were quite analogous in general; however, not all verbs had an equally high score. Verbs such as *confirm*, *protect* had a considerably lower score. Unlike subjects, the objects of the verbs differed between phishing and legitimate emails. A possible explanation is that objects play a bigger role in delivering the sender's intent, thus significantly limiting the domain. Looking at the result, most of the objects in phishing emails are *information*, *account*, and *link*; on the other hand, the objects in legitimate emails include *PC*, *receipt*, and *meeting*. They convey the main part of intention of senders respectively, but with the much larger domain in the legitimate communication.

The results of cosine similarity between all subjects and all objects is consistent with this explanation. The verb *click* had almost identical objects regardless of the emails. This is easily explainable as a number of things that are clickable is limited in normal life as well – especially, those that would be given the direction to in an email. On the other hand, the verb *access* has many more possibilities that are of interest to a person phishing for information. To sum up, the similarity scores of parse tree path and subject and object were not consistent, and their characteristics seemed to depend on verbs themselves rather than a clear-cut differentiation between phishing or legitimate categories.

The cosine similarity for all subjects and all objects between the two phishing corpora was measured. The results indicated that most verbs had similar subjects and objects between the two data. In particular, contrary to the object comparison with legitimate emails, a majority of the same objects appeared in both phishing data. Some exceptions are explained in the previous section. Based on the results, we could conclude that the patterns in phishing emails had not been substantially altered in terms of text contents. Given this evidence, it is likely that current and future phishing scams will still contain slightly modified texts in emails. This is a promising sign for our future research in that semantically analyzed subjects and objects can tighten the robustness of their features against changes in phishing emails over time. For instance, the most frequent objects for *confirm* and *update* were *identity* and *records* in Nazario phishing data, and they both changed to *information* in APWG phishing data. From the perspective of semantics, *identity, records, and information* can belong to the same domain.

The experiment presented in this paper expanded the analysis unit from words to segments of sentences, and attempted to compare the verb-centered networked features in sentences compared to the previous works only dealing with simple word matching. The next step is to find patterns in email texts in terms of word meanings, and cluster them into semantic domains. We expect this work will be able to deal with the syntactically different, but semantically identical or similar words, and thereby produce features to be generalized. The ultimate objective of our research is to determine the hidden intention of email from the computer perspective so that machines could more accurately detect phishing emails. Since overall semantic processing is considered to be a heavy artillery, we attempted a lighter version as a stepping stone in the overall detection. Our future research will focus on semantically processing emails to detect phishing attacks based on the aims of emails rather than shallow distinguishable features.